\documentclass{llncs}

\usepackage{hyperref}
\usepackage{url}

\usepackage{times}
\usepackage{latexsym}
\usepackage{amsmath}
\usepackage{amssymb}
\usepackage{bm}
\usepackage{algorithm}
\usepackage{algpseudocode}
\usepackage{xcolor,trimclip}
\usepackage{graphicx}
\usepackage{float}
\usepackage{booktabs}
\usepackage{longtable}
\usepackage{caption} 
\usepackage{array}

\begin{document}

\title{Continuous Semantic Topic Embedding Model Using Variational Autoencoder}
\titlerunning{CSTEM}  
%
\author{Namkyu Jung \and Hyeong In Choi}


%
\authorrunning{Namkyu Jung et al.} 
%
\tocauthor{Namkyu Jung, Hyeong In Choi}
\institute{Seoul National University, Gwanak-gu, Seoul 151-742, South Korea,\\
\email{ \{lep, hichoi\}@snu.ac.kr}
}

\maketitle              

\begin{abstract}
This paper proposes the continuous semantic topic embedding model (CSTEM) which finds latent topic variables in documents using continuous semantic distance function between the topics and the words by means of the variational autoencoder(VAE). The semantic distance could be represented by any symmetric bell-shaped geometric distance function on the Euclidean space, for which the Mahalanobis distance is used in this paper. In order for the semantic distance to perform more properly, we newly introduce an additional model parameter for each word to take out the global factor from this distance indicating how likely it occurs regardless of its topic. It certainly improves the problem that the Gaussian distribution which is used in previous topic model with continuous word embedding could not explain the semantic relation correctly and helps to obtain the higher topic coherence.
Through the experiments with the dataset of 20 Newsgroup, NIPS papers and CNN/Dailymail corpus, the performance of the recent state-of-the-art models is accomplished by our model as well as generating topic embedding vectors which makes possible to observe where the topic vectors are embedded with the word vectors in the real Euclidean space and how the topics are related each other semantically.
\end{abstract}

\section{Introduction}
Topic models give a probability of words appearing in a text by discovering latent topic variables which have their own distribution of words. The probabilistic Latent Semantic Indexing (pLSI) provided by \cite{Hofmann:1999:PLS:312624.312649} suggested the semantic probabilistic technique for analysis co-occurrences between words and documents and the Latent Dirichlet Allocation(LDA) by \cite{Blei:2003:LDA:944919.944937} gives us Bayesian probabilistic generative models by allocating latent topic variables, which is a generalization of pLSI. 

LDA model basically assumes that the occurrence of each vocabulary in a document is influenced by its latent topic variable which is represented as a categorical distribution over words. And these topic variables follow Dirichlet distribution parametrized by Dirichlet priors of their documents, which play role as mixing coefficients of topic distributions. The latent variables of topics and Dirichlet priors are learned from several unsupervised learning algorithms.
\cite{Blei:2003:LDA:944919.944937} solved the intractability of the posterior distribution by variational Bayesian method \cite{Jordan1999}, and \cite{Griffiths06042004} presented a Markov chain Monte Carlo algorithm(MCMC) with Gibbs sampling for inference of LDA. Although both methods had successfully inferred this model, the variational method converges fast whereas the MCMC converges more correctly because of bias-variance trade-off. And these methods have been researched a lot for the better performance and efficiency.

\cite{Porteous:2008:FCG:1401890.1401960} introduced the FastLDA algorithm which enhanced the convergence speed 8 times faster than the standard collapsed Gibbs sampler and \cite{canini2009online} proposed online inference Gibbs sampling algorithms. \cite{teh2007collapsed} proposed a collapsed variational Bayesian inference for LDA and \cite{NIPS2010_3902} had developed an online variational Bayes algorithm for it. Development for applying LDA to large-scale data \cite{newman2008distributed}, \cite{yan2009parallel}, \cite{wang2009plda} and other applications to various area \cite{Krestel:2009:LDA:1639714.1639726}, \cite{lienou2010semantic}, \cite{lukins2008source}, \cite{Maskeri:2008:MBT:1342211.1342234}, \cite{putthividhy2010topic}, \cite{Biro:2008:LDA:1451983.1451991}, \cite{Arora:2008:LDA:1390749.1390764} have been researched actively until present.

While the neural network has extended their area massively, to text processing as well \cite{miao2016neural}, very efficient inference method using neural networks called Auto-Encoding Variational Bayes (AEVB) for the intractable posterior was introduced in \cite{DBLP:journals/corr/KingmaW13}. This method straightforwardly optimize a variational lower bound estimator using standard stochastic gradient methods so that the intractable posterior can be inferred efficiently by fitting an approximate inference model. AEVB is naturally applied to the topic model, which is successfully realized by Autoencoded Variational Inference For Topic Model(AVITM) proposed in \cite{AVITM}, to which our paper mainly refer. 

Topic distribution through LDA cannot imply the correlation between the topics which is measured by the topic coherence \cite{newman2010automatic}, \cite{mimno2011optimizing}, \cite{lau2014machine}. On that count, \cite{Das2015GaussianLF} tried to introduce multivariate Gaussian distribution to the word-topic distribution with fast collapsed Gibbs sampling
algorithm based on Cholesky decompositions 
by virtue of the word representation called \texttt{word2vec}. It asserted that the method has achieved an average 275\% higher topic coherence on average. Following it, other researches using Gaussian LDA \cite{hu2012latent}, \cite{xun2017correlated} have introduced models to achieve better topic coherence and shown a state-of-the-art performance. AVITM \cite{AVITM} also showed higher topic coherence than any other before. However, these methods have not considered the influence of the word itself to the occurrence in a documents regardless of the topic if there is an assumption of continuity for the words distribution per-topic. If we think that the continuous word-topic distribution has something to do with semanticity, there supposed to be consideration of the influence of each word itself to the probability of occurrence, explained at chapter \ref{chaper_weight} in detail.

Thus, our paper proposes the model named Continuous Semantic Topic Embedding Model (CSTEM) which applies a continuous word-topic distribution through the AEVB method. We presumed that if the word-topic distribution were represented to a continuous function then it should have \textit{semantic meaning} in it, as it is placed in the \textit{semantic field} constructed on the Euclidean space. So we considered several \textit{semantic distance} functions to make the word-topic distributions and see how it fits the real text data well. Additionally, we introduce a new probabilistic model parameter which represents global weight for each word regardless of topics in the consideration of assumption that the semantic distance between the word and the topic could not give an enough explanation for the probability of words occurrence. We claim that this global parameter will fit the missing part of the continuous topic model. In order to learn this new parameter not depending on the datapoints, we apply full variational autoencoder method which is presented in the appendix of \cite{DBLP:journals/corr/KingmaW13}. Furthermore, this model with neural network generates the topic and word representations as by-products which helps us to check out their location on the Euclidean space visually.

In short, the fruits of our paper is the following.

\begin{itemize}
\item The model catches up or slightly outperforms previous topic model in accuracy and topic coherence in shorter learning time than the sampling-based model.
\item Embedding topic vectors on the Euclidean space obtained from the model gives us not only the semantic distances between topics and words but also semantic distances between topics themselves.
\item The distance function learned by the model reflects the semantic relation more realistically by introducing the global weight representing how commonly each word occurs regardless of the topic.
\end{itemize}

\section{Background}
In this paper, we use a corpus $\mathcal{D}$ containing $N$ documents $\{ \textbf{w}_1, \cdots, \textbf{w}_N \}$ consisting of vocabularies $\{ w^1, \cdots, w^V \}$ where $V$ is the size of all vocabularies. Each document $\mathbf{w}_d$ has $N_d$ words $\{w_{d1}, \cdots, w_{d,N_d}\}$ and $w_{dn}$ means the $n$-th word in the $d$-th document. Every random variable $w$ can be regarded as a one-hot encoding $V$-dimensional vector, has 1 at its index and 0 elsewhere.

\subsection{Latent Dirichlet Allocation}

We use Latent Dirichlet Allocation (LDA) model as a base generative model.
LDA assumes that for each document $d$ in $\mathcal{D}$ has a latent variable  
$\theta_d \sim \textrm{Dirichlet}(\alpha)$, a conjugate prior for the categorical distribution of topics for each words $ z \sim \textrm{Categorical}(\theta_d) $ in a document $d$, where $\alpha$ is a  parameter of Dirichlet distribution. There is a model parameter $\beta \in \mathbb{R}^{K\times V}$ representing the probability for word appearing for each topic. $\beta$ denotes
$\beta=(\beta_1, \cdots, \beta_K)$, where $K$ is a number of topics and each $\beta_k$ is a parameter of categorical distribution over the vocabulary. That is, 
$$ p(w_{dn}=w^v | z_n=k, \beta) = \beta_{kv}. $$
Under this model, by collapsing the latent variable $z$, 
the marginal likelihood of $\mathbf{w}_d$ is

\begin{align*}
p(\mathbf{w}_d|\alpha, \beta)=&\int_{\theta_d}\Big(\prod_{n=1}^{N_d} \sum_{k=1}^K
p(w_{dn}|z_n=k, \beta) p(z_n=k|\theta_d)\Big)p(\theta_d|\alpha)d\theta_d
\end{align*}

\subsection{Gaussian LDA}
In order to assume Gaussian distribution for the word-topic distribution, 
there supposed to be a mean and a covariance matrix for each topic $k$ denoting 
$\bm{\mu}_k$, $\bm{\Sigma}_k$ so that $w_{dn}|z_{dn} \sim 
\mathcal{N}(\bm{\mu}_k, \bm{\Sigma}_k)$. But since Gaussian distribution is on
the continuous domain so we need $w_{dn}$ to be a continuous real vector.
Gaussian LDA model in \cite{Das2015GaussianLF}, \cite{ijcai2017-588} 
uses word embeddings like \texttt{word2vec} so that each $w_{dn}$ could be embedded to
$\mathbf{w}_{dn}\in \mathbb{R}^W$ where $W$ is the dimension of word embeddings.
Therefore,
$$w_{dn}|z_{dn} \sim \mathcal{N}(\mathbf{w}_{dn};\bm{\mu}_{z_{dn}}, \bm{\Sigma}_{z_{dn}}) $$

\subsection{Autoencoding Variational Bayes}
When the marginal likelihood is not analytically tractable as in LDA, the variational autoencoder in 
\cite{DBLP:journals/corr/KingmaW13} is one of the methods for the inference of generative models.
Since there is only one latent variable $\theta$ by collapsing $z$, it needs a variational posterior distribution $q_{\phi}(\theta)$ with the variational parameter $\phi$. The marginal log-likelihood of the
corpus is the sum of the log-likelihood of each document $\mathbf{w}_d$, which can be represented as:
\begin{align}
\log p(\mathbf{w}_d | \alpha, \beta) = D_{KL}(q_{\phi}(\theta_d|\mathbf{w}_d) 
|| p(\theta_d|\mathbf{w}_d, \alpha, \beta)) + \mathcal{L}(\phi, \alpha, \beta;\mathbf{w}_d), \label{loglikelihood}
\end{align}
where 
\begin{align}
\mathcal{L}(\phi, \alpha, \beta;\mathbf{w}_d)
=-D_{KL}(q_{\phi}(\theta_d| \mathbf{w}_d)||p(\theta|\alpha)) + \mathbb{E}_{q_{\phi}(\theta_d|\mathbf{w}_d)}[\log p( \mathbf{w}_d | \theta_d, \beta)] \label{elbo1}
\end{align}
is called the ELBO(evidence lower bound) and $D_{KL}$ represents the KL-divergence. 
Note that the first term of the ELBO is the KL-divergence between the varational posterior and the true posterior, while the second term of it is the expected negative \textit{reconstruction error}.
\cite{AVITM} uses the Gaussian distribution as $q_{\phi}(\theta)$ to avoid the difficulty of 
reparametrization trick \cite{DBLP:journals/corr/KingmaW13} by taking advantage of the Laplace 
approximation between the Dirichlet distribution and the Gaussian distribution \cite{DBLP:journals/corr/abs-1110-4713}, so that $p(\theta|\alpha) \simeq \hat{p}(\theta|\bm{\tilde{\mu}}_{\theta}, \bm{\tilde{\Sigma}}_{\theta})=\mathcal{N}(\bm{\tilde{\mu}}_{\theta}, \bm{\tilde{\Sigma}}_{\theta})$ where 
\begin{align}
\tilde{\mu}_{\theta k}&=\log \alpha_k - \frac{1}{K}\sum_i \log \alpha_i \label{alpha_to_mu}\\
\tilde{\Sigma}_{\theta kk} &=\frac{1}{\alpha_k}\Big(1-\frac{2}{K}\Big)+\frac{1}{K^2}\sum_i \frac{1}{\alpha_i} \label{alpha_to_sigma} .
\end{align}
Note that $q_{\phi}(\theta)$ has mean $\mu_{\theta}=f_{\mu}(\mathbf{w}, \mathbf{\delta_{\mu}})$ and diagonal covariance $\Sigma_{\theta}=f_{\mathbf{\Sigma}}(\mathbf{w}, \delta_{\Sigma})$ where $f_{\mu}, f_{\Sigma}$ are the feed forward neural networks with the parameters $\delta_{\mu}, \delta_{\Sigma}$. From this variational posterior, we can compute the reconstruction error by Monte Carlo method generating sample $\theta_d=\sigma(\bm{\tilde{\mu}}_{\theta} + \bm{\tilde{\Sigma}}_{\theta}^{1/2}\epsilon)$ with sampling $\epsilon \sim \mathcal{N}(0, \mathbf{I})$. 
The word-topic distribution parameter $\beta$, which is a model parameter, would be optimized as the model converges.

\section{Continuous Semantic Topic Embedding Model}

\subsection{Semantic Distance Between Word and Topic}
Our model aims that the topic vectors and word vectors will be learned by VAE in \textit{semantic} way. In other words, if a word is frequent in a certain topic, the vectors of this word and topic should be
\textit{semantically close}. Therefore, we need a metric on the embedding space to measure how close between the topic and the word, which can be regarded as a \textit{semantic distance} between them. In this sense, Gaussian LDA
uses a density function of Gaussian distribution as a metric so that words with high probability have larger likelihood than the others, which means that it is more likely to be sampled.

However, we don't need to have a Gaussian assumption, and moreover, any function which is centralized and bell-shaped can be used as a metric because we assume that the geometric closeness means the semantic closeness. (Only symmetric function will be considered.) The simplest distance function which satisfies this condition is the Mahalanobis distance, defined as
$$ D_M(x;\mu, \Sigma) = \sqrt{(x-\mu)^T\Sigma^{-1}(x-\mu)}.$$
Note that $\mu$ is a topic embedding vector playing a role as a \textit{semantic center} in the word embedding space and $\Sigma$ as a scaling factor.
Naturally, the word-topic distribution can be defined inversely proportional to this distance like 
$x|\mu, \Sigma \sim \dfrac{1}{D_M(x;\mu, \Sigma)^2 + \epsilon}$, where $\epsilon$ is for the case of zero distance.
We can use other functions as a distance function such as the density function of Cauchy distribution, Student t-distribution and the logistic distribution. The difference between them is how much they are centralized, decided by their variances. The larger variance it has, the more likely this model could estimate words relatively far. In this paper the Mahalanobis distance will be used for the simplicity.
\begin{figure}[h]
\centering
\includegraphics[scale=0.5]{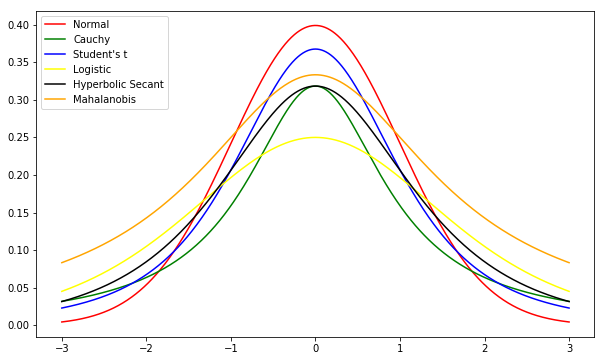}
\caption{
	Symmetric centralized distributions which can be uses as a \textit{semantic distance}. 
}
\end{figure}

\subsection{Global Weight Parameter of Words}\label{chaper_weight}
The probability of a certain word occurring in some topics is decided not only by the semantic distance between the word and the topic but also by the global tendency of occurrence of the word itself. For instance, if we consider a topic as \textit{football}, the word \textit{hat-trick} is semantically much closer than the word \textit{take} because the word \textit{take} has no specific relation with the topic \textit{football}. However we can easily estimates that the word \textit{take} will be occurring more than the word \textit{hat-trick} in a topic \textit{football} because \textit{hat-trick} is not happening much often in a football game, while the word \textit{take} will occurs frequently since it is a commonly used word no matter in which topic it is contained. Therefore the generative model should have a factor representing this weight for each vocabulary so that the model can fit the real data more accurately. Note that this factor has nothing to do with the semantical relation, thus it does not dependent on the topics.
Under the assumption that $w$ depends on $c$, the global weight parameter $c = (c_{1}, \cdots, c_{V}) \sim \textrm{Dirichlet}(\gamma)$, the joint probability of word occurring in a document would be
\begin{align}
p(w, \theta, c | \alpha, \beta, \gamma) &= p(w|\theta, \beta, c)p(\theta|\alpha)p(c|\gamma)\\
&=\sum_{k=1}^K p(w|z=k, \beta, c)p(z=k|\theta)p(\theta|\alpha)p(c|\gamma).
\end{align}
where $\gamma$ is a prior distribution of $c$.
We intend that the probability $p(w|z=k, \beta, c)$ would be proportional to $c$ for any given topic $k$ and $\beta$.
Therefore, the marginal likelihood of the model can be modified as
\begin{align}\label{marginal_lkhd}
p(\mathbf{w}_d |\alpha, \beta, \gamma)=
\int_c \int_{\theta_d} \prod_{n=1}^{N_d} \sum_{k=1}^K p(w_{dn}|z_n=k, \beta, c) p(z_n=k|\theta_d)p(\theta_d|\alpha)p(c|\gamma) d\theta_d dc.
\end{align}
Later, we will assume the variational distribution of $c$ in order to regularize it in VAE way. Therefore we will use the approximation $c|\gamma \sim \textrm{Dirichlet}(\gamma) \simeq \mathcal{N}(\bm{\tilde{\mu}}_{c}, \bm{\tilde{\Sigma}}_{c})$ in the same way as the relations (\ref{alpha_to_mu}), (\ref{alpha_to_sigma}).

\subsection{Word-Topic Distribution With Topic Embedding}
With the well-defined semantic distance metric $d(x;\mu, \Sigma)$, and the parameter $c_v$ for each word $w^v$, our model proposes
the inference of the
topic-word distribution to be
\begin{align}
p(w=w^v|z=k, \beta, c) &:=
\textrm{softmax}\Big(\frac{c_v}{d(\textbf{w}^v;\bm{\mu}^{k}_\beta, \bm{\Sigma}^{k}_\beta)^2 + \epsilon}\Big) \\
&=\frac{\exp \left\{\dfrac{c_v}{d(\textbf{w}^v;\bm{\mu}^{k}_\beta, \bm{\Sigma}^{k}_\beta)^2 + \epsilon} \right\}}{\sum_{l=1}^V \exp \left\{\dfrac{c_l}{d(\textbf{w}^l;\bm{\mu}^{k}_\beta, \bm{\Sigma}^{k}_\beta)^2 + \epsilon} \right\}} \label{def_beta},
\end{align}
where $\mathbf{w}^v=\textrm{Emb}(w^v) = (W_{\textrm{Emb}})^T w^v$ is the embedding vector of the word $w^v$
 (considered to be a one-hot encoding vector $\in\mathbb{R}^V$)
and $W_{\textrm{Emb}} \in \mathbb{R}^{V \times W}$ is the word embedding matrix, 
$\bm{\mu}^{k}_\beta, \bm{\Sigma}^{k}_\beta$ for 
$k=1, \cdots, K$  are the topic mean vectors and the topic covariance matrix for each $k$. Simply speaking, the probability of a certain word occurring is exponentially proportional to the square inverse of semantic distance $d$ and the global weight $c$.
Note that $W_{\textrm{Emb}}$ and $\bm{\mu}^{k}_\beta, \bm{\Sigma}^{k}_\beta$ are the model parameters expected to be learned
by VAE where $\beta$ represents the set of all model parameters, $\beta=(W_{\textrm{Emb}}, \bm{\mu}^{k}_\beta, \bm{\Sigma}^{k}_\beta)$, so do not confuse with LDA notation. The definition (\ref{def_beta}) can be different when the semantic distance is defined differently.
This equation makes possible to maintain the structure of LDA and get topic vectors and 
word vectors at the same time. Combining (\ref{def_beta}) with the marginal likelihood equation (\ref{marginal_lkhd}) ultimately represents the structure of generative model CSTEM. The inference of CSTEM is executed by the VAE as discussed above. Next section derives the process of inference in VAE framework.

\subsection{Variational Objective Function}
To define the variational objective function we have to maximize, let us assume the variational posteriors $q(\theta, c|\mathbf{w}) = q(\theta|\mathbf{w})q(c)$, one with the topic prior and the other with the global weight parameter, to be
\begin{align*}
\log q_{\phi}(\theta|\mathbf{w}) &= \log \mathcal{N}(\theta; \bm{\mu}_{\theta}, \bm{\Sigma}_{\theta})\\
\log q_{\phi}(c) &= \log \mathcal{N}(c; \bm{\mu}_c, \bm{\Sigma}_c)
\end{align*}
where $\bm{\mu}_c$ and $\bm{\Sigma}_c=\bm{\sigma}_c\textbf{I}$ are variational parameters and $\bm{\mu}_{\theta}, \bm{\Sigma}_{\theta}=\bm{\sigma}_{\theta}\textbf{I}$ are to be learned by the neural networks from $\textbf{w}$ as $\bm{\mu}_{\theta} = f_{\mu}(\textbf{w}, \delta_{\mu})$, $\bm{\Sigma}_{\theta}=f_{\Sigma}(\textbf{w}, \delta_{\Sigma})$.
Note that the true priors of them are Gaussian distributions
\begin{align*}
\log p(\theta|\alpha) &= \log \mathcal{N}(\theta; \bm{\tilde{\mu}}_{\theta}, \bm{\tilde{\Sigma}}_{\theta})\\
\log p(c|\gamma) &= \log \mathcal{N}(c; \bm{\tilde{\mu}}_{c}, \bm{\tilde{\Sigma}}_{c}),
\end{align*}
induced from the Laplace Approximation of the Dirichlet distribution in the same way as the equation (\ref{alpha_to_mu}), (\ref{alpha_to_sigma}).
We suggest to have variational posteriors of $\theta, c$ to have close distribution to the true priors $p(\theta), p(c)$ in order to regularize them. It has to be clear that even though the parameters of the word-topic distribution $\beta$ could be inferred in variational way like the distribution of $c$ as well if we assume the prior distribution of it, we just infer them assuming they are constants being optimized by the stochastic method for a couple of reasons; (i) not necessary to regularize each topic mean vectors and covariance vectors and (ii) to make it simple with considering the memory issue.

\begin{figure}[]
\centering
\includegraphics[scale=0.4]{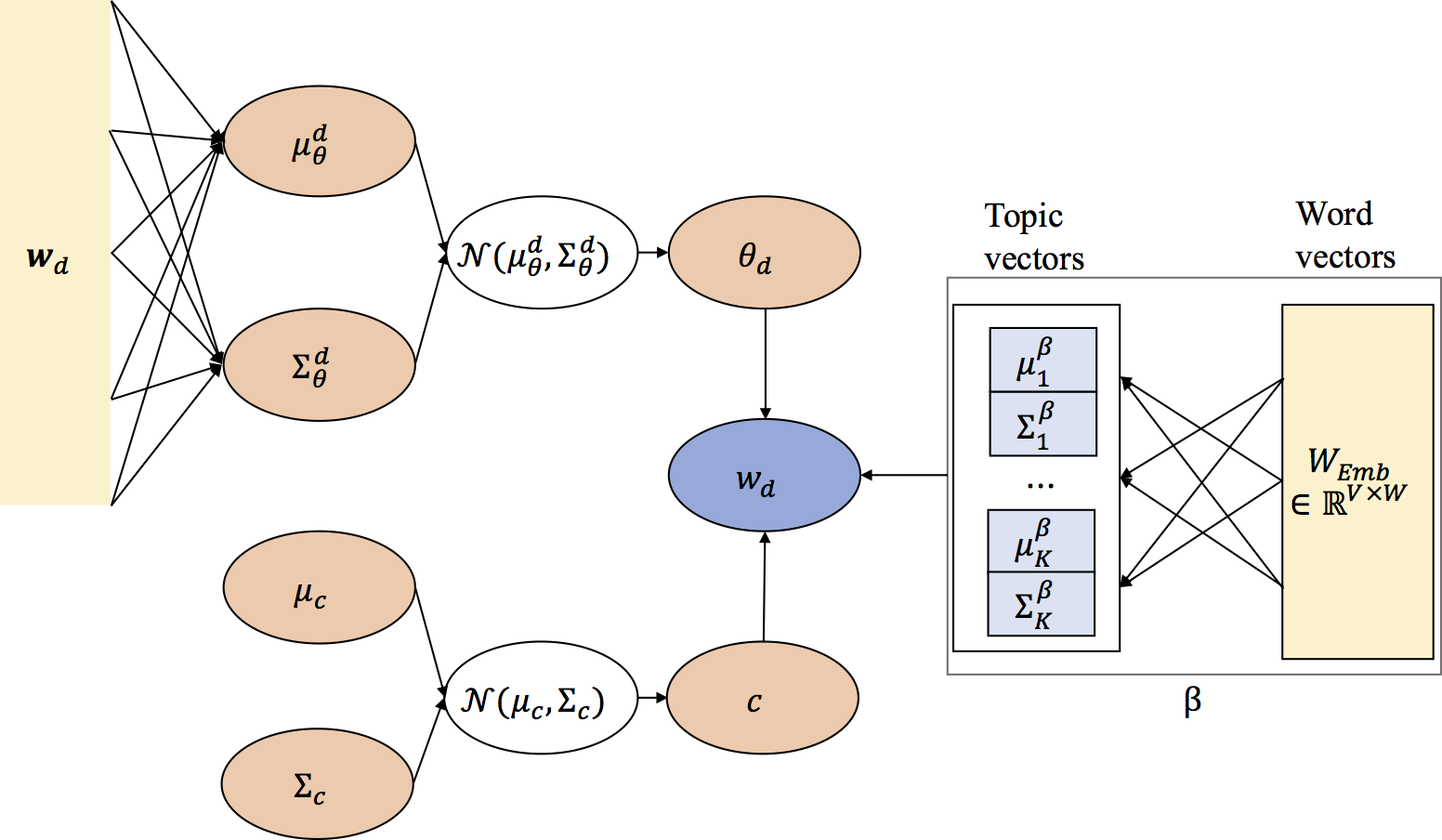}
\caption{
	The graphical model of the inference process for CSTEM. It consists of the encoder part on the top-left, the inference of variational parameter $c$ on the botton-left and the decoder part using the semantic distance on the right. 
\label{diagram}
}
\end{figure}

The marginal likelihood of the full corpus can be written as:
\begin{align*}
\log p(\mathcal{D})=D_{KL}(q_{\phi}(c) || p(c|\mathcal{D}))+\mathcal{L}(\phi;\mathcal{D})
\end{align*}
similar to (\ref{loglikelihood}) and the ELBO as:
\begin{align*}
\mathcal{L}(\phi;\mathcal{D})=
-D_{KL}(q_{\phi}(c)||p(c)) + \mathbb{E}_{q_{\phi}(c)}[\log p(\mathcal{D}| c)]
\end{align*}
similar to (\ref{elbo1}).
Since the term $\log p(D|c)$ is a sum of marginal log-likelihoods of each data points, it can be written as:
\begin{align*}
\log p(\mathcal{D}|c) &= \sum_{d=1}^N \log p(\mathbf{w}_d|\alpha, \beta, c) \\
&= \sum_{d=1}^N D_{KL}(q_{\theta_d}(c) || p(\theta_d|\textbf{w}_d))+\mathcal{L}(c,\alpha,\beta,\phi;\textbf{w}_d)
\end{align*}
where the ELBO is:
\begin{align}
\mathcal{L}(c,\alpha,\beta,\phi;\textbf{w}_d)=
-D_{KL}(q_{\phi}(\theta_d)||p(\theta_d|\alpha)) + \mathbb{E}_{q_{\phi}(\theta_d|\textbf{w}_d)}[\log p(\textbf{w}_d|\theta_d; \beta, c)].
\end{align}
Since the variational approximate posteriors are both Gaussian, we can sample it using reparametrization trick as:
\begin{align}
q_{\phi}(\theta_d|\textbf{w}_d) &\quad \text{ as }\quad \theta_d= \mu^d_{\theta} + \sigma^d_{\theta} \odot \zeta \hspace{40pt} \textrm{where} \quad \zeta \sim \mathcal{N}(0, \mathbf{I}) \label{rep_theta}\\ 
q_{\phi}(c) &\quad \text{ as }\quad c=\mu_{c} + \sigma_{c} \odot \epsilon \hspace{40pt} \textrm{where} \quad \epsilon \sim \mathcal{N}(0, \mathbf{I}), \label{rep_c}
\end{align}
where $\odot$ means an element-wise product. Thus if we set the variational parameters $\phi=(\mu_{c}, \Sigma_{c}, \delta_{\mu}, \delta_{\Sigma})$, the resulting objective function results in
\begin{align*}
\mathcal{L}(\beta, \phi; \mathcal{D}) & \simeq 
-D_{KL}(q_{\phi}(c)||p(c)) + \mathbb{E}_{q_{\phi}(c)}\Big[\sum_{d=1}^{N} \mathcal{L}(c,\alpha,\beta,\phi;\textbf{w}_d)\Big]\\
&=-D_{KL}(q_{\phi}(c)||p(c)) + \mathbb{E}_{q_{\phi}(c)} \Big[ \sum_{d=1}^{N} 
-D_{KL}(q_{\phi}(\theta_d)||p(\theta_d|\alpha)) \\ & \quad + \mathbb{E}_{q_{\phi}(\theta_d|\textbf{w}_d)}[\log p(\textbf{w}_d|\theta_d; \beta, c)]\Big]\\
&=-D_{KL}(\mathcal{N}(\bm{\mu}_c, \bm{\Sigma}_c)||\mathcal{N}(\bm{\tilde{\mu}}_{c}, \bm{\tilde{\Sigma}}_{c})) \\ & \quad + \sum_{d=1}^{N} -D_{KL}(q_{\phi}(\theta_d)||p(\theta_d|\alpha)) + \mathbb{E}_{\zeta}\Big[
\mathbb{E}_{\epsilon}[\log p(\textbf{w}_d|\theta_d; \beta, c)]\Big]\\
&= - \frac{1}{2} \Big(
tr (\bm{\tilde{\Sigma}}_{c}^{-1} \bm{\Sigma}_{c})+ (\bm{\tilde{\mu}}_{c}-\bm{\mu}_{c})^T 
\bm{\tilde{\Sigma}}_{c}^{-1} (\bm{\tilde{\mu}}_{c}-\bm{\mu}_{c}) - V + \log \frac{|\bm{\tilde{\Sigma}}_{c}|}{|\bm{\Sigma}_{c}|} \Big) \\
& \quad + \sum_{d=1}^{N} \Big[ - \frac{1}{2} \Big(
tr (\bm{\tilde{\Sigma}}_{\theta}^{-1} \bm{\Sigma}^d_{\theta})+ (\bm{\tilde{\mu}}_{\theta}-\bm{\mu}^d_{\theta})^T 
\bm{\tilde{\Sigma}}_{\theta}^{-1} (\bm{\tilde{\mu}}_{\theta}-\bm{\mu}^d_{\theta}) - K + \log \frac{|\bm{\tilde{\Sigma}}_{\theta}|}{|\bm{\Sigma}^d_{\theta}|} \Big)\\
& \qquad\qquad + \mathbb{E}_{\zeta, \epsilon} [\textbf{w}_d^T \log(
\textbf{A}(\beta,c) \sigma(\theta_d))] \Big]
\end{align*}
where $\textbf{A}(\beta,c) \in \mathbb{R}^{V \times K}$ is a matrix with $\textbf{A}(\beta,c)_{v, k} = p(w=w^v|z=k, \beta, c)$ at (\ref{def_beta}), $\sigma$ denotes the softmax function, and $\theta_d, c$ are calculated from (\ref{rep_theta}), (\ref{rep_c}) respectively. The graphical model of this inference is presented in Fig. \ref{diagram}.
Using stochastic gradient variational Bayes in \cite{DBLP:journals/corr/KingmaW13}, with minibatches of size $M<D$ and the sampling size of Monte Carlo is $L$, the estimated ELBO will be
\begin{align*}
\mathcal{L}(\beta, \phi, D) &\simeq \tilde{\mathcal{L}}^M(\beta, \phi, D)\\
&= - \frac{1}{2} \Big(
tr (\bm{\tilde{\Sigma}}_{c}^{-1} \bm{\Sigma}_{c})+ (\bm{\tilde{\mu}}_{c}-\bm{\mu}_{c})^T 
\bm{\tilde{\Sigma}}_{c}^{-1} (\bm{\tilde{\mu}}_{c}-\bm{\mu}_{c}) - V + \log \frac{|\bm{\tilde{\Sigma}}_{c}|}{|\bm{\Sigma}_{c}|} \Big) \\
&+\frac{N}{M}\sum_{d=1}^M \frac{1}{L}\sum_{l=1}^L \Big[ - \frac{1}{2} \Big(
tr (\bm{\tilde{\Sigma}}_{\theta}^{-1} \bm{\Sigma}^d_{\theta})+ (\bm{\tilde{\mu}}_{\theta}-\bm{\mu}^d_{\theta})^T 
\bm{\tilde{\Sigma}}_{\theta}^{-1} (\bm{\tilde{\mu}}_{\theta}-\bm{\mu}^d_{\theta}) - K + \log \frac{|\bm{\tilde{\Sigma}}_{\theta}|}{|\bm{\Sigma}^d_{\theta}|} \Big)\\
&\qquad \qquad+ \textbf{w}_l^T \log \big(
\textbf{A}(\beta,\bm{\mu}_c+\sigma_c\odot\epsilon^{(l)})\sigma(\bm{\mu}_{\theta}^d+\sigma_{\theta}^d \odot \zeta^{(l)})\big) \Big].
\end{align*}
The pseudocode is as following.

\begin{algorithm}[H]
   \caption{CSTEM through stochastic VAE with $L=1$}
   	\hspace*{\algorithmicindent} \textbf{Input:}
    Initialize the parameters $\beta=(\bm{\mu}_\beta$, $\bm{\Sigma}_\beta, W_{Emb})$
    and $\phi=(\mu_{c}, \Sigma_{c}=\sigma_c \textrm{I}, \delta_{\mu}, \delta_{\Sigma})$.
	\begin{algorithmic}[H]
    \Repeat
      \For{each mini-batch of size $M$ = $\{d_1, \cdots, d_M \}$}
      	\State $\zeta \gets $ Random draw from prior $\mathcal{N}(0, \textbf{I})$
        \State $\epsilon \gets $ Random draw from prior $\mathcal{N}(0, \textbf{I})$
        \State $c = \mu_c + \sigma_c \odot \zeta$
        \For{$i=1, \cdots, M$}
          \State $\mathbf{\mu}^{d_i}_\theta \gets f_\mu(\textbf{w}_{d_i}, \delta_{\mu})$, $\mathbf{\sigma}^{d_i}_\theta \gets f_\sigma(\textbf{w}_{d_i}, \delta_{\Sigma})$
          \State $\theta_{d_i} \gets \bm{\mu}^{d_i}_\theta + \bm{\sigma}^{d_i}_\theta \odot \epsilon$
        \EndFor
        \State $ \mathbf{g} \gets \nabla_{\psi, \phi}\mathcal{L}(\psi, \phi; D)$
      \EndFor
    \Until convergence of $(\psi, \phi)$
\end{algorithmic}
\hspace*{\algorithmicindent} \textbf{Output:} $\beta, \phi$
\end{algorithm}

\section{Experiments}

\subsection{Practical Issues}
When it comes to the inference of model parameters $\beta=(\bm{\mu}_\beta, \bm{\Sigma}_\beta, W_{Emb})$, assuming $\bm{\Sigma}_\beta$ to be full matrix(technically, symmetric positive semi-definite matrix) takes so many memory space that for the dimension of embedding space $d_w=\textrm{dim}(\textbf{w})$, 50 or 300 in our experiments, it takes space of $\frac{d_w(1+d_w)}{2}$, exceeding the limit of our experiment resources. Although the full matrix is certainly supposed to bring the better optimization, we have used the diagonal matrix as $\bm{\Sigma}_\beta$, which still shows good enough result, and leave it with the full matrix as a future work.

\subsection{Description}
With the model proposed above, we have carried several experiments with the corpora from 20 Newsgroups \footnote{http://qwone.com/~jason/20Newsgroups/} (from \texttt{scikit-learn} \footnote{http://scikit-learn.org/stable/datasets/twenty\_newsgroups.html}), CNN/DailyMail \footnote{http://cs.nyu.edu/~kcho/DMQA/} and NIPS papers\footnote{https://www.kaggle.com/benhamner/nips-papers/data}. Every vocabulary is trimmed for meaningless words like auxiliary verbs, prepositions or articles and singularized, lemmatized and stemmed by \texttt{nltk}\footnote{http://www.nltk.org/} and we use just 2000 most frequent words. (The more the number of words is, the better performance would be expected, but we just have done with 2000 words.) 20 Newsgroups has 9402 training documents and 6252 test documents excluding documents whose number of vocabularies is less than 30. In CNN/DailyMail, we randomly picked 10000 training documents as the training set and 2000 for the test set. NIPS has total 6560 papers and we split them into the training 70\% of them and the test 30\% of them.

We use pre-trained \texttt{word2vec}\footnote{https://code.google.com/archive/p/word2vec/} and \texttt{glove}\footnote{https://nlp.stanford.edu/projects/glove/} as initial vectors of $W_{Emb}$ for the far better performance even though the random initialization still works. Since the dimension of word representation decides the dimension of topic embedding space (embedded in to the same Euclidean space as $\mathbb{R}^{d_w}$), the dimension of embedding space is chosen as 50 and 300, the dimension of \texttt{word2vec} and \texttt{glove}. Every code is implemented in \texttt{python} via \texttt{Tensorflow} based on the code of AVITM implementation\footnote{https://github.com/akashgit/autoencoding\_vi\_for\_topic\_models}. The model is trained by ADAM optimizer \cite{kingma2014adam} with learning rate 0.0002 and the calculating machine with \texttt{NVIDIA Tesla K80}. 

Since these experiments are executed as an unsupervised learning, the standard measure of how models fit well is required and traditionally the perplexity has been used for this purpose. However this measure is recently treated that it cannot reflect the topic coherence well, thus new evaluation methods have been developed recently and we will use both the  perplexity and the topic coherence measures PMI, NPMI and $\textrm{score}_\textrm{UMASS}$ from \cite{newman2010automatic}, \cite{mimno2011optimizing}, which are calculated as the average value of every pairs of top $n$ words in topic $k$. 

We compare our model to the standard LDA with collapsed Gibbs sampling \cite{Griffiths06042004} and ProdLDA from \cite{AVITM} which outperforms the other model in its own paper. We expect similar perplexity and topic coherence to ProdLDA because we aim at not only maintaining high topic coherence but also implementing continuous structure in order to get continuous representations of topics and words on Euclidean space. All other hyperparameters are applied as same for each model with mini-batch size $M=100$ and 100 epochs are executed for ProdLDA and CSTEM, while 20 epochs for Gibbs LDA because of its long elapsed time.

\subsection{Results}
Table \ref{comparison_perplexity} shows the perplexities over models. CSTEM follows right behind ProdLDA in both the train set and the test set. LDA has very large gap between the one with train set and the one with test set. It seems that LDA overfits the dataset while the models with VAE succeed to minimize the generalization error. CSTEM is actually not expected to exceed the ProdLDA's perplexity because CSTEM has structural limit of inference word-topic distribution while ProdLDA has no limit on this distribution and inferring the weights freely. However the perplexity has nothing to do with the topic coherence \cite{lau2014machine}. Thus we have to compare the topic coherence measures between them.
\begin{table}[]
\centering
\begin{tabular}{@{}lllll@{}}
\toprule
Model   & dataset & 5       & 10      & 20      \\ \midrule
LDA     & train & 572.55  & 511.44  & 460.52  \\
        & test  & 1560.89 & 1594.22 & 1539.84 \\
ProdLDA & train & 1199.34 & 1251.17 & 1166.83 \\
        & test  & 1195.52 & 1253.52 & 1162.24 \\
CSTEM   & train & 1255.93 & 1238.78 & 1212.42 \\
        & test  & 1252.78 & 1235.76 & 1209.49 \\ \bottomrule
\end{tabular}
\caption{Comparison of perplexity between models using the 20 Newsgroup dataset. For the train set, LDA is far better than any other models but for the test set, ProdLDA is generally better than others.}
\label{comparison_perplexity}
\end{table}
Table \ref{comparison1} shows that our CSTEM model show higher topic coherence than other models when the words are used more than a certain number. It means that our model finds topic of data better in a wide range but ProdLDA finds the small number of keywords better than others. It is expected phenomenon since CSTEM learns the model structure to fit the topic-word relation generally on the Euclidean space, whereas the other model does not assume any sharing space as CSTEM but only specific weights for each topic. This structure of CSTEM therefore shows the higher performance in general aspect, not much well in local perspective. Thus this model is more suitable for the purpose of general topical analysis of datasets. The elapsed time for each model is 1167, 115, 235 seconds for collapsed Gibbs LDA, ProdLDA and CSTEM respectively. Table \ref{comparison2} provides the comparison of topic coherences of CSTEM over the topic numbers. Since there is not much difference observed over the number of topics in topic coherence, we could find topics of text in detail if we increase the number of topic without losing much topic coherence.(Appendix \ref{word-topic_dist} shows that the classification with more number of topics are still performed well but just done with detail.)
\begin{table}[H]
\centering
\begin{tabular}{@{}llllllll@{}}
\toprule
Model               & Topic coherence & 5             & 10            & 20            & 30            & 50            & 100           \\ \midrule
Collapsed Gibbs LDA & NPMI            & 0.25          & 0.21          & 0.16          & 0.15          & 0.15          & 0.14          \\
                    & PMI             & 1.08          & 0.94          & 0.67          & 0.63          & 0.66          & 0.68          \\
                    & UMASS           & 2.28          & 2.39          & 2.40          & 2.50          & 2.64          & 2.81          \\ 
ProdLDA             & NPMI            & \textbf{0.45} & \textbf{0.36} & 0.23          & 0.21          & 0.17          & 0.14          \\
                    & PMI             & \textbf{2.61} & \textbf{2.10} & \textbf{1.42} & 1.30          & 1.10          & 0.95          \\
                    & UMASS           & \textbf{4.52} & \textbf{4.56} & \textbf{4.42} & \textbf{4.37} & 4.25          & 4.20          \\
CSTEM			    & NPMI            & 0.34          & 0.30          & \textbf{0.25} & \textbf{0.23} & \textbf{0.22} & \textbf{0.19} \\
          		    & PMI             & 2.03          & 1.84          & 1.56               & \textbf{1.49} & \textbf{1.40} & \textbf{1.24} \\
                    & UMASS           & 4.31          & 4.45          & 4.35          & 4.34          & \textbf{4.32} & \textbf{4.22} \\
                    \bottomrule
\end{tabular}
\caption{Comparison of topic coherences using the 20 Newsgroup dataset. The numbers of the top row means the number of words used to evaluate the scores in each topic. The bold ones represent the highest score among the models for each experiment. The higher numbers we use in evaluation, CSTEM shows the better score than others.}
\label{comparison1}
\end{table}

\begin{table}[]
\centering
\begin{tabular}{@{}lllll@{}}
\toprule
Model & 5    & 10   & 20   & 50   \\ \midrule
NPMI  & 0.20 & 0.19 & 0.20 & 0.18 \\
PMI   & 0.62 & 0.63 & 0.70 & 0.59 \\
UMASS & 1.91 & 2.02 & 2.17 & 2.03 \\ \bottomrule
\end{tabular}
\caption{Comparison of topic coherences of CSTEM over the number of topics using NIPS dataset. The numbers of top row is the number of topics. }
\label{comparison2}
\end{table}
Table \ref{common_words} shows the words whose global weight parameter $c_v$ is in top 20 for each topic. This result implies that this model learns the global weights to be high for the more common words, not having strong topical characteristic, though there are some topic-specific words in list. Considering this, we could surmise that the global weight of the word plays two important roles; (i) the model fits the dataset more accurately by adding this free parameter and (ii) it learns semantic distance function more semantically as excluding global occurrence factor with this weight and concentrating on the word-topic relation. The second role of the global weight makes the list of the geometrically closest words of each topic (Table \ref{topic_dist1}, \ref{topic_dist2}) consist of semantically closest words without the words with high global frequency like the words in the Table \ref{common_words}. Actually the model without this global weight is learned as the topics will contain some common, not related words or finds a topic whose top words mostly consist of common words. We think that this structure of word-topic distribution filtered by this factor is what the topic model originally should pursue.

Table \ref{comparison_embedding} shows some interesting points about the influence of word embedding initialization. The model initialized with \texttt{word2vec} outperforms those with \texttt{glove} for most of cases. Moreover the higher dimension used, the better it finds topics under the same word representation model. It implies that \texttt{word2vec} is more proper model than \texttt{glove} model as well as the higher dimension of embedding space used in initialization leads to the better performance, which is easily predictable. Thus we could expect nicer results if we use better word representation.

\begin{table}[]
\centering
\begin{tabular}{@{}ll@{}}
\toprule
Dataset      & Common words                                                                      \\ \midrule
20 Newsgroup & day, length, hint, pain, review, give, category, cry, travel, disagree       \\
CNN          & hear, floor, admit, resolve, greatest, someone, fell, deliver, structure, surgery \\
DailyMail    & lead, better, bin, record, heat, imagine, false, recovery, feature, become        \\
NIPS         & utility, performance, key, index, contour, explore, circle, play, iid, randomize  \\ \bottomrule
\end{tabular}
\caption{List of words which have 10 highest global weight factor for each dataset.}
\label{common_words}
\end{table}

\begin{figure}[H]
\centering
\includegraphics[scale=0.3]{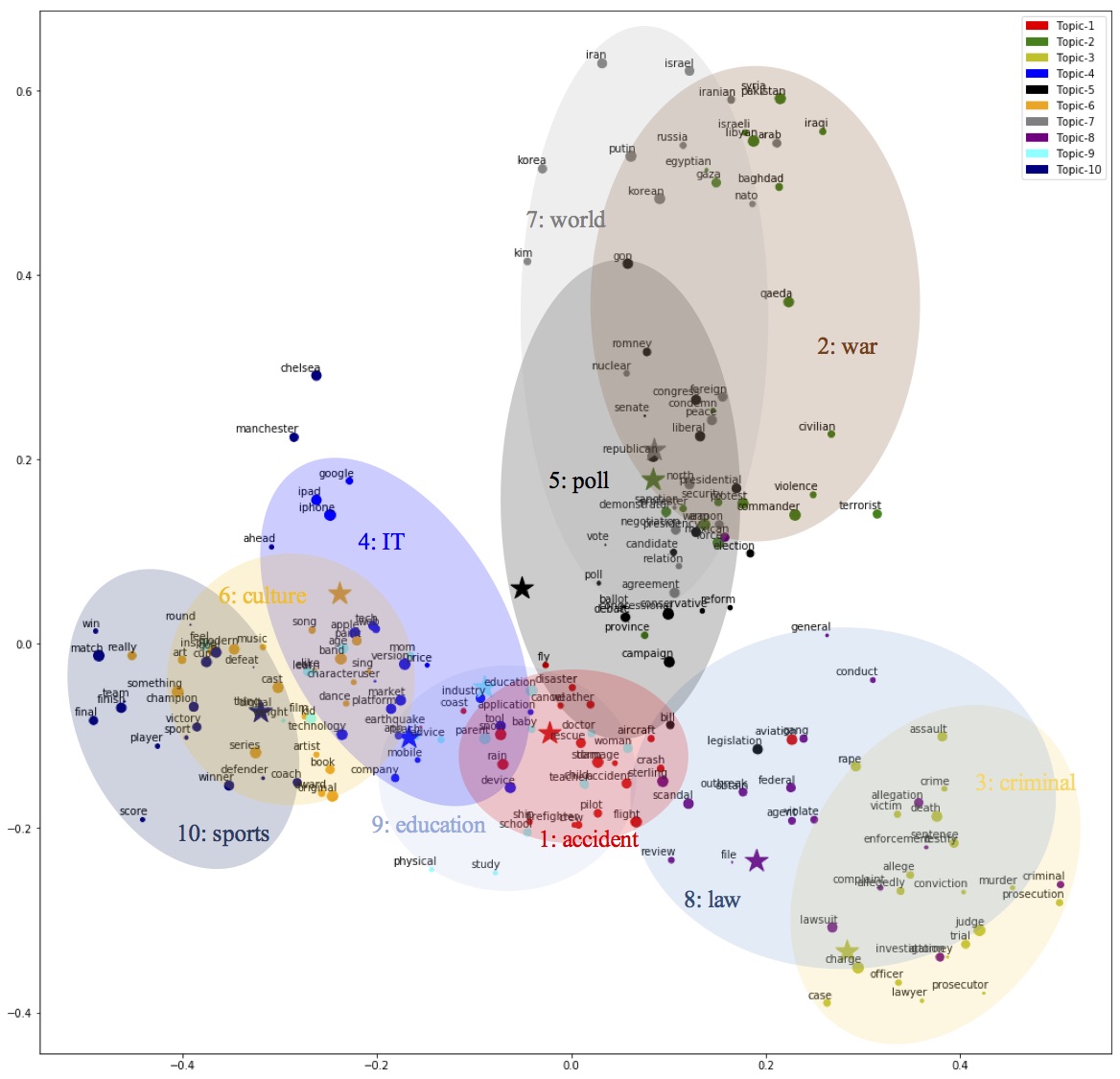}
\caption{
Word and topic embedding graph of CNN news corpus in $\mathbb{R}^2$ with dimension reduction by PCA. The dot represents the position of each word embedding vector and the star represents the position of each topic embedding vector $\mu_\beta$. Plotted words consist of top 20 words for each topic allowing repetition over topics. The radius of each word is proportional to its global weight $c_i$. Every location on $\mathbb{R}^2$ can be distorted during the dimension reduction.
}
\label{word_map}
\end{figure}

\begin{table}[]
\centering
\begin{tabular}{@{}llllllll@{}}
\toprule
Model         & Topic coherence & 5    & 10   & 20   & 30   & 50   & 100  \\ \midrule
\texttt{word2vec}(300) & NPMI            & 0.32 & 0.30 & 0.23 & 0.21 & 0.19 & 0.17 \\
              & PMI             & 1.91 & 1.88 & 1.43 & 1.31 & 1.24 & 1.14 \\
              & UMASS           & 4.14 & 4.32 & 4.23 & 4.18 & 4.20 & 4.18 \\
\texttt{glove}(300)    & NPMI            & 0.26 & 0.23 & 0.19 & 0.18 & 0.18 & 0.16 \\
              & PMI             & 1.66 & 1.51 & 1.20 & 1.18 & 1.15 & 1.07 \\
              & UMASS           & 4.23 & 4.28 & 4.20 & 4.15 & 4.12 & 4.11 \\
\texttt{glove}(50)     & NPMI            & 0.27 & 0.22 & 0.15 & 0.13 & 0.13 & 0.13 \\
              & PMI             & 1.70 & 1.40 & 0.90 & 0.81 & 0.78 & 0.81 \\
              & UMASS           & 4.00 & 3.99 & 3.86 & 3.89 & 3.89 & 3.91 \\ \bottomrule
\end{tabular}
\caption{Comparison of topic coherence over the word representation models used as a initialization of word embedding matrix $W_\textrm{Emb}$. The numbers in the parenthesis is the dimension of its embedding space.}
\label{comparison_embedding}
\end{table}
Above all, this model generates the embedding vectors for every word and topic on the Euclidean space. We expect that the word vectors are located close if they are semantically related and the topic vectors are close to the word vectors which are belong to them. Thus we plot words and topics from the experiment using CNN news dataset with 10 topics through the dimension reduction via PCA\cite{wold1987principal}. (Detail words distribution is shown at the Table \ref{topic_dist1} with 20 topics.) Even though PCA distorts the embedding vectors since the dimension of vectors are reduced from 300 to 2, observing Figure \ref{word_map}, the words map in which top 20 words for each topic are plotted, we can check that words and topics are gathered semantically as expected. And the word vectors are re-optimized according to the per-topic relations in comparison to the initial word embedding vectors.

Besides, by virtue of the topic embedding to the Euclidean space, we can \textit{see} that the topics with semantical similarity are gathered geometrically by observing their embedding vector through PCA. In Fig. \ref{word_map}, the vector of topic named \textit{war}, for instance, is close to that of topic named \textit{world}. Actually they have similar meaning in this corpus since the nations' name in the topic \textit{world} is commonly mentioned with the articles about \textit{war}, such as \textit{israel}, \textit{korea} and \textit{iran}. We can find more topical relations like the one between \textit{law} and \textit{criminal}, and the one between \textit{sports} and \textit{culture}. We could conclude that the semantic distance we have learned actually plays role as the semantic distance between topics themselves. Since the distance is defined continuously, we could generate custom topic we want infinitely by picking some vector on the embedding space. This is the main achievement of this paper and we make the semantic distance have only semantic meaning by adding global weight factors.

\section{Conclusion and Future Works}
In this paper, we proposed a new topic model CSTEM which is a generative model allocating latent topic variables by learning semantic distance function and global word weights. It assumes that the probability of words occurrence weight for a certain topic is exponentially proportional to the inverse of the properly defined semantic distance and the global occurrence of each word regardless of the topic it belongs to. This structure makes possible to generate the continuous embedding vectors of words and topics on the Euclidean space as well as fit the relation between the topic variables and the words.

The model has two latent variables to infer, $c$ and $\theta_d$, which can be inferred by the VAE method. Unlike $\theta_d$ requiring maximizing likelihood for each data point, $c$ does not depend on each data point, which leads to the maximizing likelihood for the full dataset. We executed optimization using the reparametrization trick and the ELBO we derived with Monte Carlo method. 

This model results in the reasonable topic coherence as well as the perplexity which measures the goodness of topic models. Additionally we checked out that this model is appropriate for the purpose of fitting text data with global analysis since this model outperform other models when we measure the topic coherence with more words. Learning time the model takes is longer than ProdLDA in some degree but still much shorter than the inference based on sampling method, collapsed Gibbs LDA.

This paper has not done with experiments for comparing to other recent topic model like CTM(\cite{lafferty2006correlated}, \cite{ijcai2017-588}), Gaussian LDA\cite{Das2015GaussianLF}, CGTM\cite{xun2017correlated} because of not enough resources. This should be done for the more concrete verification. However this paper still means something since it has achieved the better performance than LDA and ProdLDA with very short time and the continuous embeddings in addition.
And it lacks of experiments over the effect of the choice of semantic distance function other than Mahalanobis function like Gaussian distribution or Cauchy distribution. The further study of the semantic distance would be meaningful. As discussed above, we assume the covariance matrix (metric matrix) of each topic to be diagonal matrix due to the memory issue. Better environment is provided, full matrix will give us higher accuracy, though it will be much slower.
Additionally the research over the further applications using the generated embedding vectors by this model should be done in the future.

%
%

\bibliography{cstem}

\begin{thebibliography}{10}

\bibitem{Hofmann:1999:PLS:312624.312649}
Hofmann, T.:
\newblock Probabilistic latent semantic indexing.
\newblock In: Proceedings of the 22Nd Annual International ACM SIGIR Conference
  on Research and Development in Information Retrieval. SIGIR '99, New York,
  NY, USA, ACM (1999)  50--57

\bibitem{Blei:2003:LDA:944919.944937}
Blei, D.M., Ng, A.Y., Jordan, M.I.:
\newblock Latent dirichlet allocation.
\newblock J. Mach. Learn. Res. \textbf{3} (March 2003)  993--1022

\bibitem{Jordan1999}
Jordan, M.I., Ghahramani, Z., Jaakkola, T.S., Saul, L.K.:
\newblock An introduction to variational methods for graphical models.
\newblock Machine Learning \textbf{37}(2) (Nov 1999)  183--233

\bibitem{Griffiths06042004}
Griffiths, T.L., Steyvers, M.:
\newblock Finding scientific topics.
\newblock Proceedings of the National Academy of Sciences \textbf{101}(suppl 1)
  (2004)  5228--5235

\bibitem{Porteous:2008:FCG:1401890.1401960}
Porteous, I., Newman, D., Ihler, A., Asuncion, A., Smyth, P., Welling, M.:
\newblock Fast collapsed gibbs sampling for latent dirichlet allocation.
\newblock In: Proceedings of the 14th ACM SIGKDD International Conference on
  Knowledge Discovery and Data Mining. KDD '08, New York, NY, USA, ACM (2008)
  569--577

\bibitem{canini2009online}
Canini, K., Shi, L., Griffiths, T.:
\newblock Online inference of topics with latent dirichlet allocation.
\newblock In: Artificial Intelligence and Statistics. (2009)  65--72

\bibitem{teh2007collapsed}
Teh, Y.W., Newman, D., Welling, M.:
\newblock A collapsed variational bayesian inference algorithm for latent
  dirichlet allocation.
\newblock In: Advances in neural information processing systems. (2007)
  1353--1360

\bibitem{NIPS2010_3902}
Hoffman, M., Bach, F.R., Blei, D.M.:
\newblock Online learning for latent dirichlet allocation.
\newblock In Lafferty, J.D., Williams, C.K.I., Shawe-Taylor, J., Zemel, R.S.,
  Culotta, A., eds.: Advances in Neural Information Processing Systems 23.
\newblock Curran Associates, Inc. (2010)  856--864

\bibitem{newman2008distributed}
Newman, D., Smyth, P., Welling, M., Asuncion, A.U.:
\newblock Distributed inference for latent dirichlet allocation.
\newblock In: Advances in neural information processing systems. (2008)
  1081--1088

\bibitem{yan2009parallel}
Yan, F., Xu, N., Qi, Y.:
\newblock Parallel inference for latent dirichlet allocation on graphics
  processing units.
\newblock In: Advances in Neural Information Processing Systems. (2009)
  2134--2142

\bibitem{wang2009plda}
Wang, Y., Bai, H., Stanton, M., Chen, W.Y., Chang, E.Y.:
\newblock Plda: Parallel latent dirichlet allocation for large-scale
  applications.
\newblock AAIM \textbf{9} (2009)  301--314

\bibitem{Krestel:2009:LDA:1639714.1639726}
Krestel, R., Fankhauser, P., Nejdl, W.:
\newblock Latent dirichlet allocation for tag recommendation.
\newblock In: Proceedings of the Third ACM Conference on Recommender Systems.
  RecSys '09, New York, NY, USA, ACM (2009)  61--68

\bibitem{lienou2010semantic}
Lienou, M., Maitre, H., Datcu, M.:
\newblock Semantic annotation of satellite images using latent dirichlet
  allocation.
\newblock IEEE Geoscience and Remote Sensing Letters \textbf{7}(1) (2010)
  28--32

\bibitem{lukins2008source}
Lukins, S.K., Kraft, N.A., Etzkorn, L.H.:
\newblock Source code retrieval for bug localization using latent dirichlet
  allocation.
\newblock In: Reverse Engineering, 2008. WCRE'08. 15th Working Conference on,
  IEEE (2008)  155--164

\bibitem{Maskeri:2008:MBT:1342211.1342234}
Maskeri, G., Sarkar, S., Heafield, K.:
\newblock Mining business topics in source code using latent dirichlet
  allocation.
\newblock In: Proceedings of the 1st India Software Engineering Conference.
  ISEC '08, New York, NY, USA, ACM (2008)  113--120

\bibitem{putthividhy2010topic}
Putthividhy, D., Attias, H.T., Nagarajan, S.S.:
\newblock Topic regression multi-modal latent dirichlet allocation for image
  annotation.
\newblock In: Computer Vision and Pattern Recognition (CVPR), 2010 IEEE
  Conference on, IEEE (2010)  3408--3415

\bibitem{Biro:2008:LDA:1451983.1451991}
B\'{\i}r\'{o}, I., Szab\'{o}, J., Bencz\'{u}r, A.A.:
\newblock Latent dirichlet allocation in web spam filtering.
\newblock In: Proceedings of the 4th International Workshop on Adversarial
  Information Retrieval on the Web. AIRWeb '08, New York, NY, USA, ACM (2008)
  29--32

\bibitem{Arora:2008:LDA:1390749.1390764}
Arora, R., Ravindran, B.:
\newblock Latent dirichlet allocation based multi-document summarization.
\newblock In: Proceedings of the Second Workshop on Analytics for Noisy
  Unstructured Text Data. AND '08, New York, NY, USA, ACM (2008)  91--97

\bibitem{miao2016neural}
Miao, Y., Yu, L., Blunsom, P.:
\newblock Neural variational inference for text processing.
\newblock In: International Conference on Machine Learning. (2016)  1727--1736

\bibitem{DBLP:journals/corr/KingmaW13}
Kingma, D.P., Welling, M.:
\newblock Auto-encoding variational bayes.
\newblock CoRR \textbf{abs/1312.6114} (2013)

\bibitem{AVITM}
{Srivastava}, A., {Sutton}, C.:
\newblock {Autoencoding Variational Inference For Topic Models}.
\newblock ArXiv e-prints (March 2017)

\bibitem{newman2010automatic}
Newman, D., Lau, J.H., Grieser, K., Baldwin, T.:
\newblock Automatic evaluation of topic coherence.
\newblock In: Human Language Technologies: The 2010 Annual Conference of the
  North American Chapter of the Association for Computational Linguistics,
  Association for Computational Linguistics (2010)  100--108

\bibitem{mimno2011optimizing}
Mimno, D., Wallach, H.M., Talley, E., Leenders, M., McCallum, A.:
\newblock Optimizing semantic coherence in topic models.
\newblock In: Proceedings of the conference on empirical methods in natural
  language processing, Association for Computational Linguistics (2011)
  262--272

\bibitem{lau2014machine}
Lau, J.H., Newman, D., Baldwin, T.:
\newblock Machine reading tea leaves: Automatically evaluating topic coherence
  and topic model quality.
\newblock In: EACL. (2014)  530--539

\bibitem{Das2015GaussianLF}
Das, R., Zaheer, M., Dyer, C.:
\newblock Gaussian lda for topic models with word embeddings.
\newblock In: ACL. (2015)

\bibitem{hu2012latent}
Hu, P., Liu, W., Jiang, W., Yang, Z.:
\newblock Latent topic model based on gaussian-lda for audio retrieval.
\newblock In: Chinese Conference on Pattern Recognition, Springer (2012)
  556--563

\bibitem{xun2017correlated}
Xun, G., Li, Y., Zhao, W.X., Gao, J., Zhang, A.:
\newblock A correlated topic model using word embeddings.
\newblock In: Proceedings of the 26th International Joint Conference on
  Artificial Intelligence.[doi> 10.24963/ijcai. 2017/588]. (2017)

\bibitem{ijcai2017-588}
Xun, G., Li, Y., Zhao, W.X., Gao, J., Zhang, A.:
\newblock A correlated topic model using word embeddings.
\newblock In: Proceedings of the Twenty-Sixth International Joint Conference on
  Artificial Intelligence, {IJCAI-17}. (2017)  4207--4213

\bibitem{DBLP:journals/corr/abs-1110-4713}
Hennig, P., Stern, D.H., Herbrich, R., Graepel, T.:
\newblock Kernel topic models.
\newblock CoRR \textbf{abs/1110.4713} (2011)

\bibitem{kingma2014adam}
Kingma, D., Ba, J.:
\newblock Adam: A method for stochastic optimization.
\newblock arXiv preprint arXiv:1412.6980 (2014)

\bibitem{wold1987principal}
Wold, S., Esbensen, K., Geladi, P.:
\newblock Principal component analysis.
\newblock Chemometrics and intelligent laboratory systems \textbf{2}(1-3)
  (1987)  37--52

\bibitem{lafferty2006correlated}
Lafferty, J.D., Blei, D.M.:
\newblock Correlated topic models.
\newblock In: Advances in neural information processing systems. (2006)
  147--154

\end{thebibliography}
\bibliographystyle{splncs}

\appendix
\section{Word-Topic Distributions} \label{word-topic_dist}
This section shows the semantically closest words for each topic of datasets.
The name in the second column is the estimated keyword representing its topic by the author. The order of words is the inverse order of semantic distance.
\begin{longtable}{p{0.12\textwidth}p{0.18\textwidth}p{0.7\textwidth}}

\caption{Words distribution for 20 topics with the CNN news dataset generated from CSTEM.}
\label{topic_dist1}
\endfirsthead
\endhead
\hline
Topic        &Name      & Words                                                                                                                                                                                             \\ \midrule
Common words &          & hear, floor, admit, resolve, greatest, someone, fell, deliver, structure, propose, always, oppose, full, unite, request, range, tell, fund, trust, lay                                            \\ \hline
Topic 1      & Criminal & murder, testify, testimony, jackson, lawyer, jury, witness, father, judge, sheriff, daughter, death, custody, prosecutor, son, evidence, attorney, victim, case, boy                              \\ \hline
Topic 2      & Culture  & movie, character, film, laugh, music, actor, really, audience, fun, love, guy, book, comic, pop, writer, episode, artist, stuff, song, album                                                      \\ \hline
Topic 3      & People   & love, kid, remember, life, learn, feel, yes, parent, young, page, sex, woman, transcript, please, older, husband, mom, realize, thing, daily                                                      \\ \hline
Topic 4      & Mid-asia & iran, iranian, regime, foreign, sanction, democracy, nuclear, military, opposition, ally, negotiation, iraq, arab, secretary, weapon, relation, peace, egypt, nation, syria                       \\ \hline
Topic 5      & War      & force, condemn, attack, protest, threat, security, afghanistan, amid, demonstration, minister, intelligence, ambassador, baghdad, military, ethnic, official, tension, prime, personnel, activist \\ \hline
Topic 6      & Sport    & athlete, oscar, award, sport, debut, team, armstrong, season, entertainment, winner, championship, olympic, series, game, bowl, fan, star, player, talent, actor                                  \\ \hline
Topic 7      & Ship     & vessel, ship, helicopter, pirate, navy, rescue, crew, boat, port, somalia, transport, blast, injure, marine, hostage, sea, guard, japanese, accident, coast                                       \\ \hline
Topic 8      & Soccer   & champion, tournament, final, score, match, league, title, striker, team, championship, goal, cup, club, win, minute, third, winner, finish, fourth, straight                                      \\ \hline
Topic 9      & Drug \newline \&Crime    & suspect, officer, arrest, authority, incident, allegedly, custody, mexican, gunman, gang, victim, murder, mexico, warrant, marine, affiliate, drug, investigate, investigator, assault            \\ \hline
Topic 10     & World    & isis, syrian, qaeda, rebel, attack, korean, force, civilian, syria, military, militant, pakistan, baghdad, humanitarian, afghanistan, iraqi, troop, libya, fighter, ukraine                       \\ \hline
Topic 11     & Aircraft & plane, aircraft, flight, airport, jet, pilot, crash, passenger, airline, aviation, crew, accident, fly, air, safety, board, transport, transportation, route, engine                              \\ \hline
Topic 12     & Space    & space, nasa, planet, earth, market, price, project, scientist, science, surface, global, climate, production, company, business, environmental, environment, mission, per, waste                  \\ \hline
Topic 13     & Natural\newline Disaster & flood, storm, weather, rain, water, emergency, snow, hurricane, wind, damage, river, disaster, inch, temperature, earthquake, coast, resident, ice, rescue, firefighter                           \\ \hline
Topic 14     & Medical  & patient, disease, treatment, doctor, cancer, study, infection, health, weight, care, medicine, brain, healthy, researcher, virus, surgery, baby, eat, research, illness                           \\ \hline
Topic 15     & IT       & mobile, app, ipad, device, apple, iphone, user, content, google, digital, consumer, technology, facebook, application, platform, company, web, tech, internet, feature                            \\ \hline
Topic 16     & Election & democrat, gop, senate, candidate, democratic, presidential, romney, voter, republican, vote, bill, barack, obama, congress, poll, reform, primary, mccain, ballot, conservative                   \\ \hline
Topic 17     & Legal    & court, violate, lawsuit, file, legal, judge, appeal, justice, attorney, prosecutor, prosecution, conviction, supreme, defendant, allegation, case, document, sterling, federal, lawyer            \\ \hline
Topic 18     & Tour     & museum, hotel, restaurant, beach, resort, visitor, art, guest, ride, festival, mountain, explore, tourist, paint, tree, unique, park, cruise, spot, shop                                          \\ \hline
Topic 19     & Terror   & mohammed, detain, condemn, attack, bomb, extremist, raid, arabia, saudi, protester, demonstrator, commander, embassy, abu, ali, qaeda, bin, ministry, soldier, afghan                             \\ \hline
Topic 20     & Trial    & allege, murder, jail, allegedly, conviction, charge, plead, prosecutor, warrant, count, convict, sentence, prison, judge, custody, crime, arrest, accuse, assault, obtain                    \\      \hline

\end{longtable}
\vspace{5cm}

\begin{longtable}{p{0.12\textwidth}p{0.18\textwidth}p{0.7\textwidth}}

\caption{Words distribution for 10 topics with the NIPS dataset generated from CSTEM.}
\label{topic_dist2}
\endfirsthead
\endhead
\hline
Topic        & Name                   & Words                                                                                                                                                                                                     \\ \hline
Common words &                        & utility, performance, key, index, explore, circle, play, highest, iid, randomize, capability, round, divergence, measurement, diag, consider, constrain, real-time, slower, gaussian                      \\ \hline
Topic 1      & Neural network         & network, capacity, net, gate, input, output, neural, hardware, feedforward, internal, ed, phoneme, sigmoid, unit, memory, activation, character, weight, digital, associative                             \\ \hline
Topic 2      & Bayesian               & latent, multinomial, variational, topic, sampler, bayesian, likelihood, posterior, prior, gaussian, nonparametric, predictive, mixture, document, infer, corpus, distribution, poisson, model, proportion \\ \hline
Topic 3      & CNN                    & deep, object, category, annotation, feature, image, video, cnn, semantic, discriminative, detection, attribute, descriptor, pose, box, patch, train, representation, layer, convolution,                  \\ \hline
Topic 4      & Multi-armed Bandit     & round, regret, adversary, risk, bind, bandit, loss, arm, learner, sup, online, mistake, prove, budget, price, lemma, inequality, upper, proof, colt                                                       \\ \hline
Topic 5      & Nerve\newline system           & synaptic, plasticity, neuroscience, neuronal, activity, modulation, synapse, inhibitory, inhibition, membrane, cell, sensory, fire, response, stimuli, cortex, cortical, neuron, auditory, spike          \\ \hline
Topic 6      & Convex\newline optimization    & lasso, tensor, completion, convergence, matrix, recovery, siam, entry, singular, norm, dual, minimization, convexity, convex, penalty, optimization, sketch, min, primal, thresholding                    \\ \hline
Topic 7      & Image recognition      & texture, image, patch, color, region, object, saliency, frame, pixel, ica, shape, contour, surface, detect, vision, resolution, localization, wavelet, segment, track                                     \\ \hline
Topic 8      & Clustering             & hash, kernel, manifold, distance, cluster, eigenvalue, dimensionality, spectral, formulation, sch, point, nearest, projection, preserve, cut, similarity, inner, reduction, pairwise, neighbor            \\ \hline
Topic 9      & Reinforcement learning & action, agent, policy, plan, state, reinforcement, reward, episode, execute, bellman, observable, game, option, skill, exploration, environment, trajectory, horizon, history, discount                   \\ \hline
Topic 10     & Graph                  & graph, node, graphical, undirected, marginal, community, propagation, vertex, partition, tree, clique, message, parent, exact, variable, dag, edge, pairwise, causal, child                               \\ \hline
\end{longtable}

\end{document}